\documentclass[]{nature_mod}
\usepackage{amsmath,amsfonts}

\usepackage{txfonts}
\usepackage[english]{}
\usepackage[normalem]{ulem}
\usepackage{xr}
\usepackage{graphicx}
\usepackage{tocloft}
\usepackage{rotating}
\usepackage{booktabs}
\usepackage[table]{xcolor}
\usepackage{multirow}
\usepackage{adjustbox}
\usepackage{verbatim}
\usepackage{amsmath,amssymb}
\usepackage{algorithm}
\usepackage{algpseudocode}
\usepackage[hidelinks]{hyperref}
\usepackage[numbers]{natbib}

\definecolor{tableShade}{HTML}{F1F5FA} 
\definecolor{MyGray}{rgb}{0.95,0.95,0.95}

\makeatletter 
\newcount\SOUL@minus
\makeatother  

\title{Deep Contrastive Learning for Feature Alignment: Insights from Housing-Household Relationship Inference}

\author{Xiao Qian$^{1}$, Shangjia Dong$^{2}$ \& Rachel Davidson$^{3}$}

\begin{document}

\maketitle

\begin{affiliations}
 \item Graduate Research Assistant, Department of Civil and Environmental Engineering, University of Delaware, Newark, DE 19716. USA.
 \item Corresponding author, Assistant Professor, Department of Civil and Environmental Engineering, University of Delaware, Newark, DE 19716. USA. (sjdong@udel.edu)
 \item Professor, Department of Civil and Environmental Engineering, University of Delaware, Newark, DE 19716. USA.
\end{affiliations}

\begin{abstract}
\textbf{Abstract} Housing and household characteristics are key determinants of social and economic well-being, yet our understanding of their interrelationships remains limited. This study addresses this knowledge gap by developing a deep contrastive learning (DCL) model to infer housing-household relationships using the American Community Survey (ACS) Public Use Microdata Sample (PUMS). More broadly, the proposed model is suitable for a class of problems where the goal is to learn joint relationships between two distinct entities without explicitly labeled ground truth data. Our proposed dual-encoder DCL approach leverages co-occurrence patterns in PUMS and introduces a bisect K-means clustering method to overcome the absence of ground truth labels. The dual-encoder DCL architecture is designed to handle the semantic differences between housing (building) and household (people) features while mitigating noise introduced by clustering. To validate the model, we generate a synthetic ground truth dataset and conduct comprehensive evaluations. The model further demonstrates its superior performance in capturing housing-household relationships in Delaware compared to state-of-the-art methods. A transferability test in North Carolina confirms its generalizability across diverse sociodemographic and geographic contexts. Finally, the post-hoc explainable AI analysis using SHAP values reveals that tenure status and mortgage information play a more significant role in housing-household matching than traditionally emphasized factors such as the number of persons and rooms.
\end{abstract}

\section{Introduction}

Housing represents one of the most significant assets individuals own, serving as the foundation for people's daily activities and broader urban dynamics. Researchers have created synthetic populations \cite{farooq2013simulation, borysov2019generate, qian2024deep, kotelnikov2023tabddpm} to enable assessments at the household and individual levels. Understanding the characteristics of households living in various types of housing units is critical for capturing the intricacies of urban life. However, our current understanding of the relationship between housing and households remains limited. This knowledge gap hinders the creation of a comprehensive joint housing-household inventory that accurately reflects household-level disaster impacts. Consequently, it also impedes the development of effective disaster intervention policies, such as targeted relief and recovery efforts.
While previous studies have sought to develop integrated housing and household inventories \cite{rosenheim2021integration, ye2024enhancing, harada2017projecting}, they primarily emphasize population density, building capacity, or sociodemographic distribution. They often neglect the explicit linkage between households and housing units, preventing us from identifying which households are more likely to reside in specific types of housing. This missing linkage limits our ability to construct a truly integrated inventory for understanding households' interaction with the built environment and assessing disaster impacts on different populations. To address this knowledge gap, this research aims to explicitly model the housing-household relationship, enabling a more accurate analysis of housing occupancy patterns and their implications for disaster resilience and urban planning.

\begin{figure}[!ht]
    \centering
    \includegraphics[width=\linewidth]{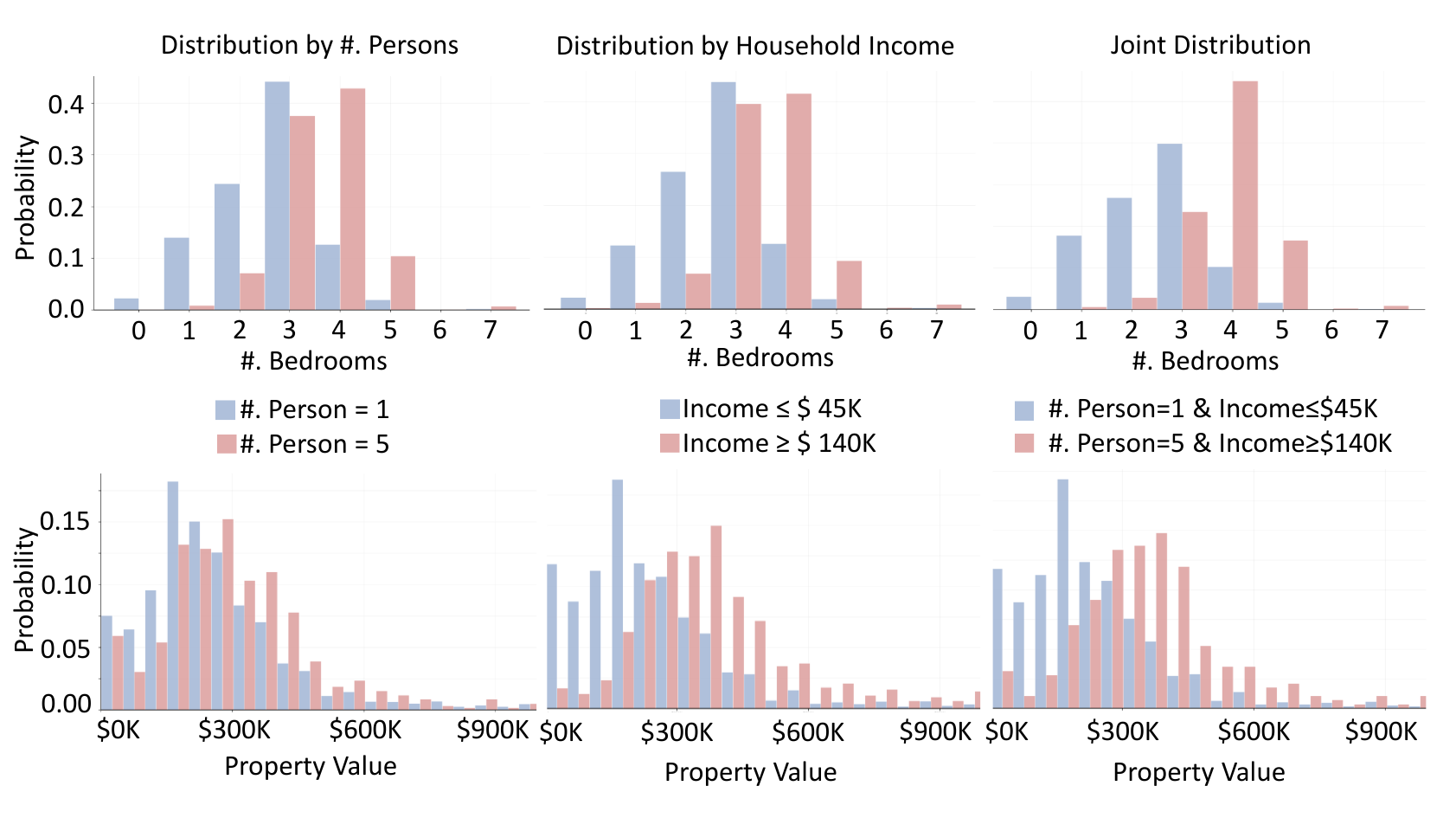}
    \caption{Microdata table illustration and descriptive statistics}
    \label{fig:desriptive_stats}
\end{figure}

The American Community Survey (ACS) Public Use Microdata Sample (PUMS) \cite{us_census_acs}, hereafter referred to as microdata, serves as a valuable resource for analyzing the relationship between households and the housing units they occupy. However, due to privacy considerations and other constraints, ACS releases only a limited 1-5\% sample of the actual population within Public Use Microdata Areas (PUMAs). Household characteristics primarily include sociodemographic information such as income, age, education, and race, while housing unit attributes describe features like the number of bedrooms, mortgage or rent payments, and property value. Figure \ref{fig:desriptive_stats} illustrates the microdata structure and some descriptive statistics, showing, for example, how households of 5 individuals can reside in units with a range of numbers of bedrooms and property values. Similarly, households with incomes $\geq \$140$k may occupy diverse types of housing units. When considering multiple sociodemographic factors—such as household size alongside income—the distribution of suitable housing options shifts. This relationship is complex and cannot be captured solely through linear models or simple rules, highlighting the intricate patterns between household demographics and housing characteristics.

Modeling the joint housing-household relationship is inherently complex, requiring advancements in data processing techniques and feature alignment methods. These challenges are not unique to housing-household relationship modeling; rather, they represent a broader class of tabular data integration problems, where ground truth linkages between two distinct datasets are missing, and only a limited number of positive co-occurring instances are available, such as pairing user profiles with product listings \cite{mcauley2015inferring}, matching patient records with medical treatments \cite{hripcsak2013next}, or aligning job applicants with job openings \cite{malinowski2006matching}. This research not only generates new knowledge on housing-household relationships but also drives broader innovations in information fusion techniques. 

To illustrate the technical advances in a concrete and relatable manner, we use housing-household relationship learning as a representative case of joint tabular data relationship modeling. However, it is important to note that the proposed methods extend beyond this specific application and can be widely applied to similar challenges in other domains.

\textbf{Challenges} Modeling housing-household relationships presents several key challenges due to inherent data limitations.
\begin{enumerate}
\vspace{-4pt}
    \item \textit{Data with positive pairs only}: microdata only provides household sociodemographics and housing unit features that co-occur within individual records. We lack explicitly labeled “negative” data points that indicate which types of households are unlikely to reside in specific types of housing. Unobserved pairs are not necessarily negative, as these may exist in reality but are simply absent from the sample. The absence of explicitly labeled negative data calls for innovative data engineering and learning techniques to ensure accurate and reliable relationship modeling.  \vspace{-6pt}  
    \item \textit{Many-to-many matching possibility}: each housing unit could be suitable for multiple households, but the microdata only records a single pairing. An alternative is to group households and housing units into different clusters and extract the many-to-many pairing information. While this broader perspective allows us to account for various housing options, it also introduces potential false positives (or noise in computer science terms) during clustering, as it implicitly assumes households may live in a variety of housing types, even if this relationship is not explicitly recorded in the microdata. Addressing this requires a clustering method capable of fine-grained grouping and a robust and noise-aware learning approach. \vspace{-6pt}  
    \item \textit{Binary matching instead of matching degree}: the co-occurrence of housing unit and household in a microdata record is a binary indicator of a household's residence choice, but the selected unit may not necessarily represent the household’s ideal living preference. Housing-household matching, therefore, requires a more nuanced approach. Ideally, we aim to provide a degree of match and rank housing-household feature alignment rather than assuming all possible units are equal likely matches. This calls for a modeling approach that can learn from diverse co-occurrences and assign a matching score to each housing-household pair.  \vspace{-6pt} 
    \item \textit{Matching subjects consist of two distinct entities}: household sociodemographics and housing unit features, representing people and buildings, respectively. Applying a standard feature alignment pipeline that assumes they share the same feature space would result in inaccurate matching. Addressing this challenge requires us to develop a specialized feature-learning pipeline that captures the unique yet interconnected nature of housing-household relationships. \vspace{-6pt}  
    \item \textit{Lack of a labeled test set}: the microdata provides only a single observed match for each household and its corresponding housing unit, and we lack a labeled dataset with negative housing-household pairs. Additionally, inferred or clustered pairs—where each household could theoretically match with multiple housing units—are plausible but not definitive. This necessitates a comprehensive and accurately labeled test set to effectively evaluate model performance.
\end{enumerate}

\textbf{Contribution} In this research, we present a deep contrastive learning (DCL) approach to model the relationship between housing unit features and household attributes. The main technical contributions of this work are as follows:
\begin{itemize}
\vspace{-4pt}
    \item We utilized the co-occurrence of housing and household data as a pretext task to address the challenge of limited positive pair data and lack of negative pair data (Challenge 1). \vspace{-6pt}  
    \item We introduced a clustering method (Section \ref{sec:bisect-cluster}) to categorize households and housing units into distinct types (Challenge 2), facilitating feature alignment between households and housing units. This grouping approach also enhances the pool of positive and negative pairs available for modeling training (Challenge 1). \vspace{-6pt}  
    \item We developed a DCL framework with dual encoders (Section \ref{sec:model-architecture}) to address the challenges of noisy data introduced by clustering (Challenge 2) and the distinct feature spaces of housing and households (Challenge 4). This framework also enables nuanced feature alignment, capturing degrees of match rather than binary outcomes (Challenge 3). Additionally, we implemented a sigmoid contrastive learning loss to account for the many-to-many relationships between households and housing units (Challenge 2). \vspace{-6pt}  
    \item To ensure robust and accurate model evaluation in the absence of a fully labeled dataset (Challenge 5), we created a synthetic ground truth to validate model performance and compare it to the state-of-the-art models prior to applying it to the microdata for learning the housing-household relationship. (Section \ref{sec:theoretical_performance}). 
\end{itemize}

\section{Literature Review}
Recognizing the importance of a joint housing and household inventory for disaster management and urban planning, several researchers have contributed to the housing-household joining effort. Harada and Murata \cite{harada2017projecting} used geospatial data to allocate households to housing units based on building specifications, such as building types and locations. Rosenheim et al. \cite{rosenheim2021integration} developed a more granular joint synthetic household and housing inventory by expanding and linking different datasets through random or rule-based processes for assigning households to housing units. Ye et al. \cite{ye2024enhancing} integrated LiDAR (Light Detection and Ranging) data, Points of Interest (POI), and quadratic programming techniques to create a population-building inventory. 

These studies illustrate the progression of joint household and housing unit modeling, evolving from random allocations to data-driven approaches. However, they also reveal a persistent challenge and research gap: the lack of housing-household relationships that capture the interaction between household characteristics and housing unit features, which is critical for advancing accurate disaster impact research. As noted by Aerts et al. \cite{aerts2018integrating}, individual and family characteristics play a significant role in decision-making processes, influencing housing choices. Ignoring these correlations undermines the accuracy and validity of subsequent research. Thus, developing methods that can accurately model housing-household relationships is essential. However, matching tabular information such as household characteristics and housing unit features faces many data and methodological challenges, as detailed earlier. 

Capturing relationships among heterogeneous datasets, commonly termed schema matching \cite{nachouki2008multi, johnston2008web}, is an active research area. Current table-matching methods largely focus on identifying content similarity between tables. For instance, the Valentine system \cite{koutras2021valentine} is an open-source experimental suite with a user-friendly graphical user interface (GUI) that is designed for large-scale schema matching. Structure-aware Bidirectional Encoder Representations from Transformers (StruBERT) \cite{trabelsi2022strubert} introduces structure- and context-aware features by integrating structural and textual information, leveraging deep contextualized language models like BERT to capture semantic similarities between tables. Similarly, Schema Matching Using Generative Tags and Hybrid Features (SMUTF) \cite{zhang2024smutf} approach combines rule-based feature engineering, pre-trained language models, and generative large language models to perform large-scale schema matching, excelling in cross-domain tasks. While these methods have advanced table-matching techniques, they mainly focus on matching similar content (e.g., Merging duplicate records in databases when two entries share overlapping attributes), whereas feature alignment involves interpreting and transforming different data types (e.g., household attributes and housing unit attributes) to reveal meaningful relationships, which is the core challenge of our research.

Recent advances in information fusion have demonstrated remarkable progress in integrating heterogeneous data sources for enhanced predictive capabilities. In biomedical applications, deep learning-based fusion methods have shown effectiveness in combining medical imaging, biomarkers, and clinical data to improve diagnostic accuracy and treatment planning \cite{zitnik2019machine, duan2024deep, zhao2024review}. In cancer research, fusion methods have successfully integrated imaging and omics data to enhance prognosis prediction and treatment response assessment \cite{lu2024privacy}.
Urban computing has similarly benefited from cross-domain data fusion techniques, particularly in integrating geographical, traffic, social media, and environmental data \cite{zou2025deep}.

While existing information fusion approaches primarily focus on aggregating heterogeneous data sources for enhanced prediction \cite{mai2023learning}, our method takes a distinct approach by learning the inherent alignment patterns from co-occurring data pairs. We leverage the co-occurrence relationships in housing-household microdata to learn a generalizable feature alignment model. This approach enables us to extend beyond the training dataset and match previously unseen housing-household pairs, providing a valuable tool for researchers in social sciences, urban computing, and disaster management. The learned feature alignment model effectively captures the complex interplay between housing and household characteristics, facilitating more accurate housing allocation analysis and disaster impact assessment.

Recent advancements in feature alignment have leveraged developments in learning-based methods from computer vision and multimodal data fusion, often referred to as Cross-modal Retrieval Methods \cite{cao2022image}. For example, the CLIP (Contrastive Language-Image Pre-training) method \cite{radford2021learning} pioneered a transformative approach by training on large-scale datasets of image-text pairs, addressing limitations in labeled data for visual recognition tasks. Building on this, ALIGN (A Large-scale ImaGe and Noisy-text embedding) \cite{jia2021scaling} scaled training data to billions of image-text pairs, achieving significant improvements in zero-shot classification. BLIP (Bootstrapping Language-Image Pre-training) \cite{li2022blip} introduced a unified framework for vision-language pre-training, incorporating captioning and filtering techniques to manage noisy web-crawled data. Other notable work includes ALBEF (Align Before Fuse) \cite{li2021align}, which used contrastive loss to align image and text representations, and UNITER (Universal Image-Text Representation) \cite{chen2020uniter}, which proposed a unified architecture for diverse vision-language tasks. These methodologies have significantly advanced cross-modal retrieval and representation learning. 

While cross-modal retrieval methods hold great promise, they are not directly suited to our objectives. These methods focus on matching data from different modalities that describe the same subject, such as pairing an image of a cat with its textual description. In contrast, our goal is to match inherently distinct features within tabular data—specifically, housing unit attributes and household characteristics. While these features pertain to different domains, one describing physical structures and the other human subjects, they share latent connections. This distinction highlights unique challenges and opportunities. While we can draw inspiration from the contrastive learning techniques employed in cross-modal retrieval, these methods must be adapted to align two different features within a single modality (tabular data). Our approach aims to capture the complex, multi-dimensional relationships between housing and household features, bridging a gap that traditional cross-modal techniques fail to address.

Given the data challenges we face, self-supervised learning with pretext tasks, where models learn from auxiliary tasks without requiring labeled data, presents a promising alternative. Pretext tasks allow models to learn from unlabeled data by solving contrived objectives that reveal inherent data patterns, with the learned representations transferable to downstream tasks. For example, BERT’s Next Sentence Prediction (NSP) task trains models to predict whether one sentence logically follows another, enabling an understanding of how two segments of discourse are connected to each other, either logically or structurally \cite{devlin2018bert}. However, RoBERTa \cite{liu2019roberta} showed that removing NSP and instead training on longer, contiguous text sequences improves downstream performance. In tabular data feature alignment, SubTab introduces pretext tasks like Reconstruction, Contrastive Learning, and Feature Vector Distance, which effectively capture tabular representations \cite{ucar2021subtab}. 

While these methods focus on general-purpose representations, our approach is tailored specifically to housing-household matching relationships. First, we use a task-specific architecture: our CLIP-inspired model processes household and housing features separately, simplifying the computational complexity from $O(N^2)$ to $O(N)$ for matching tasks. Second, we formulate a task-specific objective: we employ a single contrastive learning task optimized for housing-household alignment rather than multiple generic pretext tasks. This specialized design results in superior performance, robustness, and computational efficiency, enabling our model to address the unique challenges of housing-household matching more effectively than existing general-purpose methods, as shown in Section \ref{sec:results}. 

\section{Methodology}

We frame the learning of joint relationships between housing units and households as a feature alignment problem, with the goal of mapping them in a shared space so the compatible match can be pulled closer and incompatible ones will be distanced. We represent household-level features as $h_i \in H$, and housing unit-level features are defined as $u_j \in U$, where $H$ and $U$ are sub-tables in the microdata, capturing household and housing unit information, respectively, and $i$ and $j$ denote the row indices within the table. When $i=j$, we say that $h_i$ and $u_j$ co-occur and form a housing-household pair. 

The goal is to train an alignment function, $f_\theta(h_i, u_j)$, which maps housing unit features and household sociodemographics from their distinct spaces to a shared feature space, producing an alignment score that quantifies the degree of housing and household feature matching. At an ideal level, a well-aligned housing-household pair should exhibit relationships reflective of real-world correlations. For instance, high-income households are more likely to be matched with high-value housing units, while larger households may correlate with units offering additional vehicles or greater space.

However, these relationships are complex, displaying multivariate and nonlinear dependencies that are not easily captured by simple rules or linear models. To address this complexity, we propose a deep learning-based approach to model these high-dimensional nonlinear relationships, learning sophisticated mappings between household and housing unit features to effectively characterize their intricate dependencies. 

\subsection{Housing-Household co-occurrence in microdata}
In supervised learning, models are trained by minimizing the error between their predictions and ground-truth labels. However, in this case, we only have data on a subset of households with verified housing-household pairings, lacking a fully labeled dataset with both positive (matching) and negative (non-matching) pairs. When labeled data is limited, such as in this study, where only positive pairs are available in the microdata, self-supervised learning with pretext tasks provides an effective solution \cite{jing2020self}. This approach eliminates the need for an explicitly labeled dataset \cite{liu2021self}, as the labels are derived from the data itself such as data co-occurrence. Through the pretext task, the model learns representations that can then be transferred to downstream tasks, such as classification or regression.

\begin{figure}[!ht]
  \centering
  \includegraphics[width=0.99\textwidth]{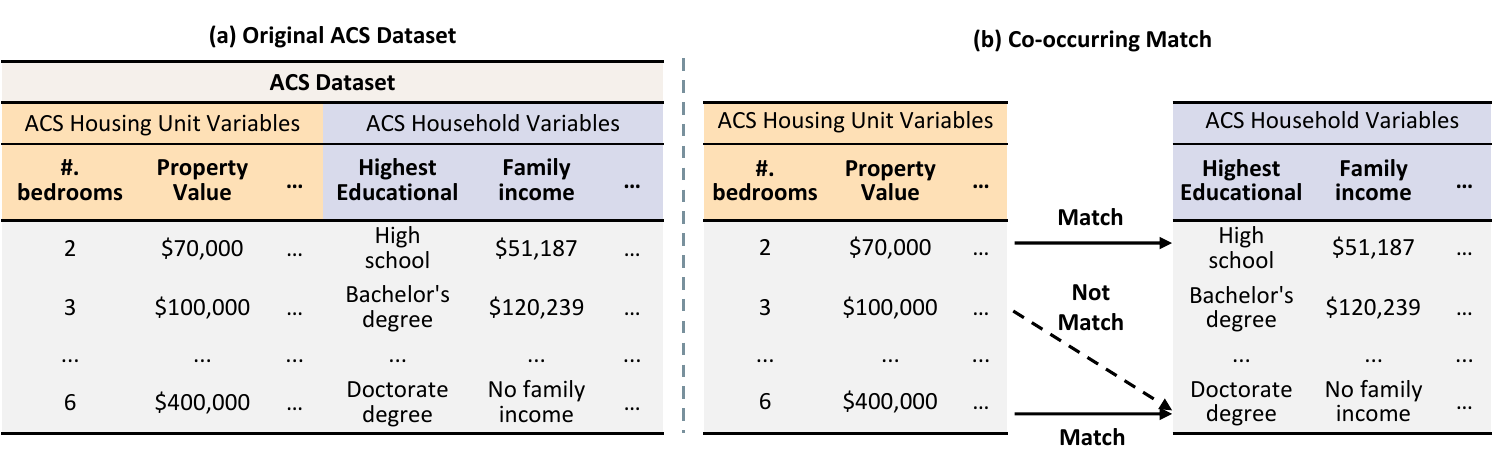}
  \caption{Illustration of the co-occurring pretext task using the ACS dataset.}
  \label{fig:co-occurring}
\end{figure}

In our study, we treat household sociodemographics and housing unit features that co-occur in the same microdata entry as positive sample pairs. The goal of our pretext task is to match housing units and households based on their co-occurrence in the microdata, which forms our pretext task. This task makes use of the matching information within the microdata and provides a learning objective that aligns closely with our main goal: to learn a relationship that enables accurate matching between compatible housing units and households. Figure \ref{fig:co-occurring} visualizes this concept using the ACS microdata. Features from the same record in ACS microdata are treated as matches, while features from different records are treated as non-matches.

Using co-occurrence as a pretext task presents several key challenges. First, due to the lack of explicit negative labels, we need to generate negative pairs for modeling. We assume that households and housing units from different rows in the microdata are incompatible, hereafter referred to as pseudo-labels \cite{jing2020self}. While the generated negative samples facilitate the training, it also introduces potential false negatives. Because the microdata only represents a sample of the entire population, households and housing units deemed "incompatible" within the sample may actually be compatible pairs in the broader population. This assumption introduces noise that requires a robust model capable of learning the true relationship between households and housing units despite these false negatives. Co-occurrence in the microdata is a binary indicator of a household's residence choice, but not all suitable housing units are equally appealing. The housing-household feature alignment task should assess the degree of matching. Third, the model must generalize beyond the microdata to match unseen households and housing units accurately. 

To address these challenges, our approach includes strategies to mitigate bias and noise in the co-occurrence-based pretext task. Specifically, in Section \ref{sec:bisect-cluster}, we introduce a noise-aware preprocessing procedure that minimizes data noise impact. Sections \ref{sec:model-architecture} details our use of a contrastive learning framework and loss function, which improve the model’s capacity for robust representation learning amidst noise. 

\begin{figure}[!ht]
  \centering
  \includegraphics[width=\textwidth]{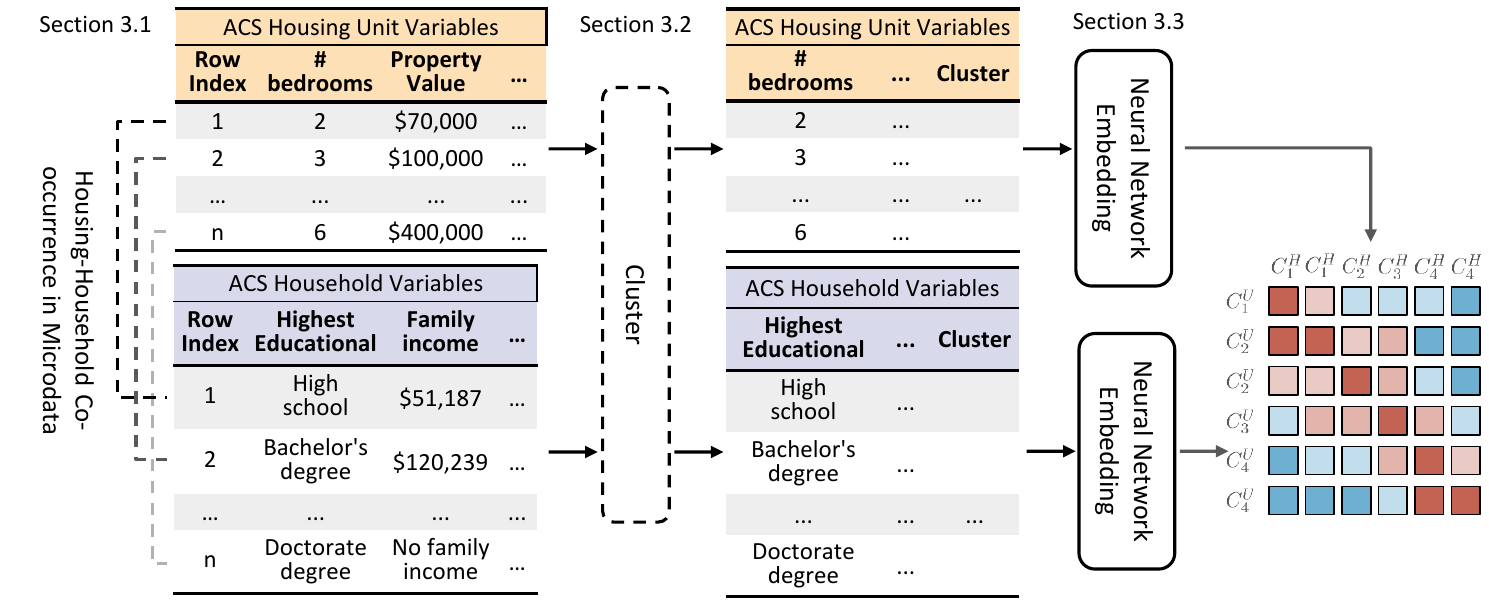}
  \caption{Joint Housing-household relationship learning pipeline}
  \label{fig:pipeline}
\end{figure}

Figure \ref{fig:pipeline} shows our proposed pipeline. The pipeline first preprocesses the microdata, and then feeds it into a dual-encoder neural network architecture. This architecture comprises separate encoders for housing unit characteristics and household sociodemographic features, which enables domain-specific learning for each feature type. During training, the two networks work collaboratively to project household and housing unit features into a shared latent space. In this space, semantically similar housing-household pairs are positioned closer together, allowing the model to capture complex, non-linear relationships and efficiently match household and housing features.

\subsection{Bisecting K-Means clustering of households and housing units}\label{sec:bisect-cluster}

The co-occurrence between individual housing units and household demographics in microdata provides a preliminary understanding of the housing-household relationship. However, microdata does not fully capture the complexity, as a single household could theoretically live in various housing units, but ultimately can only occupy one at a time. Similarly, a housing unit might be suitable for multiple households. To address this, a more useful approach is to explore what types of households are best suited for specific types of housing units.

One basic approach to classify households would be to categorize based on all possible demographic and housing attribute combinations. For instance, income or age could be divided into discrete ranges, and their combinations enumerated. However, this method is inherently arbitrary and unlikely to represent an optimal grouping of households and housing units. More critically, such simplistic combinatorial approaches treat each category as a discrete and independent entity, failing to capture relationships or distances between different categorical items. For example, a household with an income of \$49,000 would be considered entirely distinct from one with an income of \$51,000 if they fall into separate income brackets. In reality, their socioeconomic statuses are very close when income is viewed as a continuous variable. This approach fragments the feature space into isolated categorical combinations, creating "islands" that hinder the establishment of meaningful relationships between groups.

Instead, we employ a bisecting K-Means clustering approach. The process begins with feature standardization—numerical features such as household income and housing value are normalized using z-scores, while categorical features like education level and housing type are one-hot encoded. This ensures features of different scales are weighted equally in the clustering. 

Next, we apply the Bisecting K-Means method to cluster households. Unlike traditional K-Means, which partitions the dataset into k clusters in one step, Bisecting K-Means takes a hierarchical, divisive approach. It iteratively splits clusters with the largest sum of squared errors (SSE) into two sub-clusters, continuing this process until the desired number of clusters is reached. This approach provides more refined groupings, allowing us to better capture variations in housing unit and household compatibility.

\begin{figure}[!ht]
  \centering
  \includegraphics[width=0.99\textwidth]{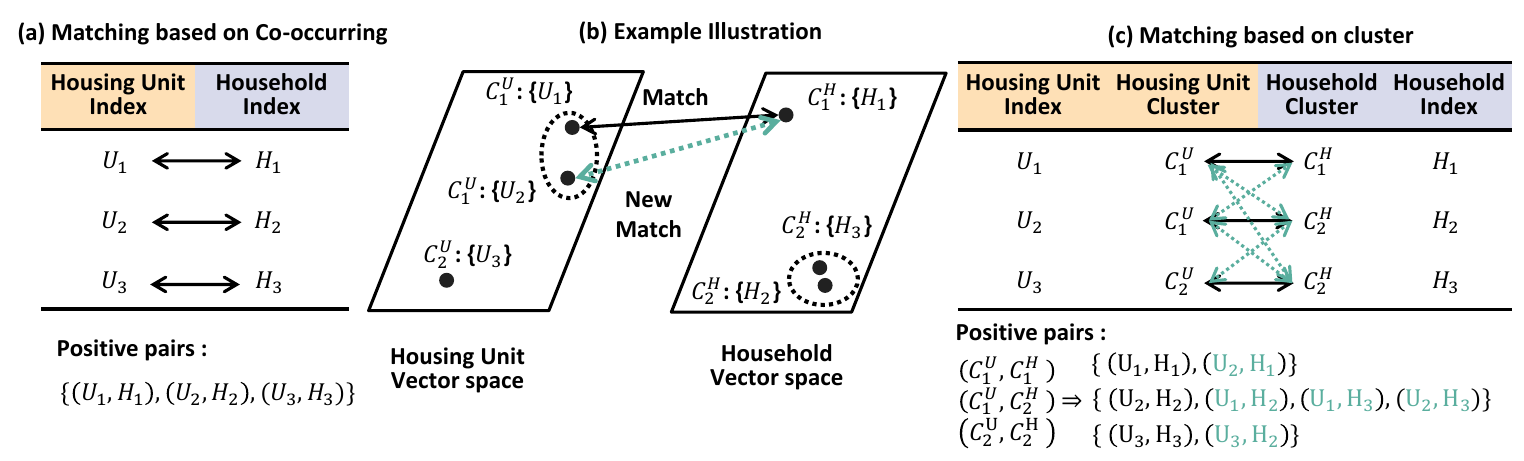}
    \caption{Illustration of the motivation and process of bisecting K-Means clustering.}
  \label{fig:cluster}
\end{figure}

In microdata, housing-household relationships are initially represented through simple one-to-one co-occurrence, as illustrated in Figure \ref{fig:cluster}(a), where each housing unit is matched with a corresponding household (e.g., ${U_1-H_1, U_2-H_2, U_3-H_3}$). However, this straightforward matching approach is overly restrictive and fails to account for similarities between housing units and households. As shown in Figure \ref{fig:cluster}(b), if two housing units (e.g., $U_1$ and $U_2$) are highly similar in the feature space, they should also be compatible with similar households. For instance, if $U_1$ matches with $H_1$, then $U_2$ should be considered a potential match for $H_1$. To systematically implement this insight, we propose a cluster-based co-occurring strategy (Figure \ref{fig:cluster}(c)). Housing units and households are clustered separately based on their feature similarities. By identifying clusters (e.g., $C_1^U$ for housing units and $C_2^H$ for households), we expand the matching relationships beyond the original co-occurrence pairs. For example, when $U_1$ and $U_2$ belong to the same cluster $C_1^U$, and $H_2$ and $H_3$ belong to cluster $C_2^H$, additional pairs such as $U_1-H_2$, $U_1-H_3$, and $U_2-H_3$ can be derived alongside the original pair $U_2-H_2$. These expanded housing-household pairs serve as augmented labels for model training, enabling the model to learn more generalized and robust matching patterns that extend beyond the limited scope of direct co-occurrence.

\begin{figure}[!ht]
  \centering
  \includegraphics[width=\textwidth]{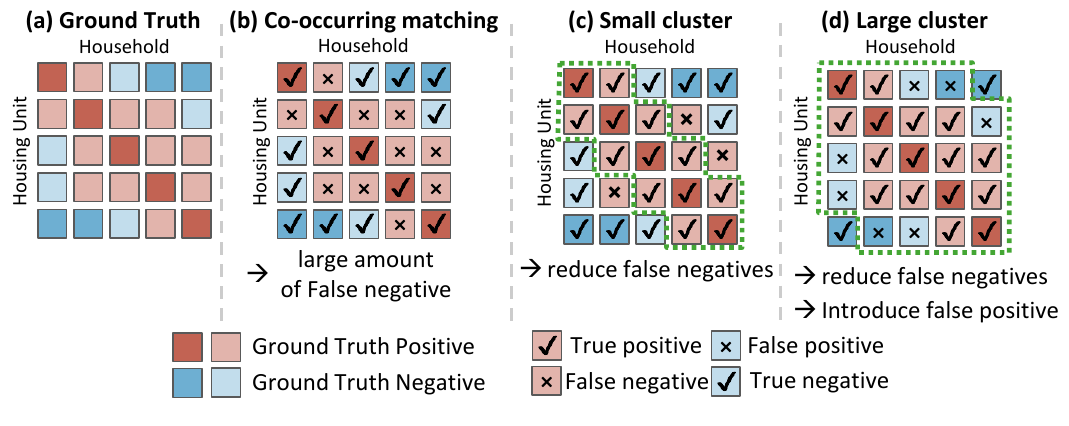}
  \caption{Impact of clustering granularity on label matching quality}
  \label{fig:clustering_error}
\end{figure}

While clustering helps capture nuanced relationships between different types of households and housing units and generates additional data for model training, it can also introduce errors. 
In the ideal scenario depicted in Figure \ref{fig:clustering_error}(a), a ground truth dataset assigns each housing-household pair a matching score represented by different shades of color, with positive samples highlighted in red and negative samples in blue. However, microdata co-occurrence, as illustrated in Figure \ref{fig:clustering_error}(b), only records observed matches along the diagonal, leaving many potentially valid matches unaccounted for. Treating all non-co-occurring pairs as negatives risks introducing false negatives, as some unobserved pairs may still be valid matches outside of microdata.

Clustering addresses this limitation by generating new housing-household pairs, capturing some of the missing valid matches. Yet, clustering creates a new challenge, as it may introduce false positives by creating housing-household pairs that are not truly compatible. The balance between true and false positives depends on the chosen cluster size. A moderate cluster size (Figure \ref{fig:clustering_error}(c)) reduces false negatives by grouping similar samples together, keeping false positives to a manageable level. Conversely, overly aggressive clustering (Figure \ref{fig:clustering_error}(d)) significantly lowers false negatives but introduces many false positives by forcing dissimilar samples into the same cluster. The impact of cluster size on data quality and model performance is critical. We investigate the impact of cluster size through sensitivity analysis in Section \ref{sec:sensitivity}.

\subsection{Deep contrastive learning (DCL) model architecture}\label{sec:model-architecture}

Joining housing units and households can be formulated as a feature alignment problem. However, due to the lack of an explicitly labeled dataset (with both positive and negative pairs), a self-supervised learning approach—contrastive learning—becomes an effective solution for this task. Contrastive learning is well-suited for learning discriminative features that can draw related housing-household pairs closer while pushing unrelated pairs farther apart. Additionally, this approach is robust to the class imbalance inherent in our dataset, where only a limited subset of co-occurring housing and household records are captured through clustering, while most other housing-household pairs are considered negative (i.e., pseudo-labels). Contrastive learning has also proven to excel in representation learning and matching for unseen data, a critical advantage since our constructed training set cannot fully capture the numerous housing-household combinations. 

\begin{figure}[!ht]
  \centering
  \includegraphics[width=\textwidth]{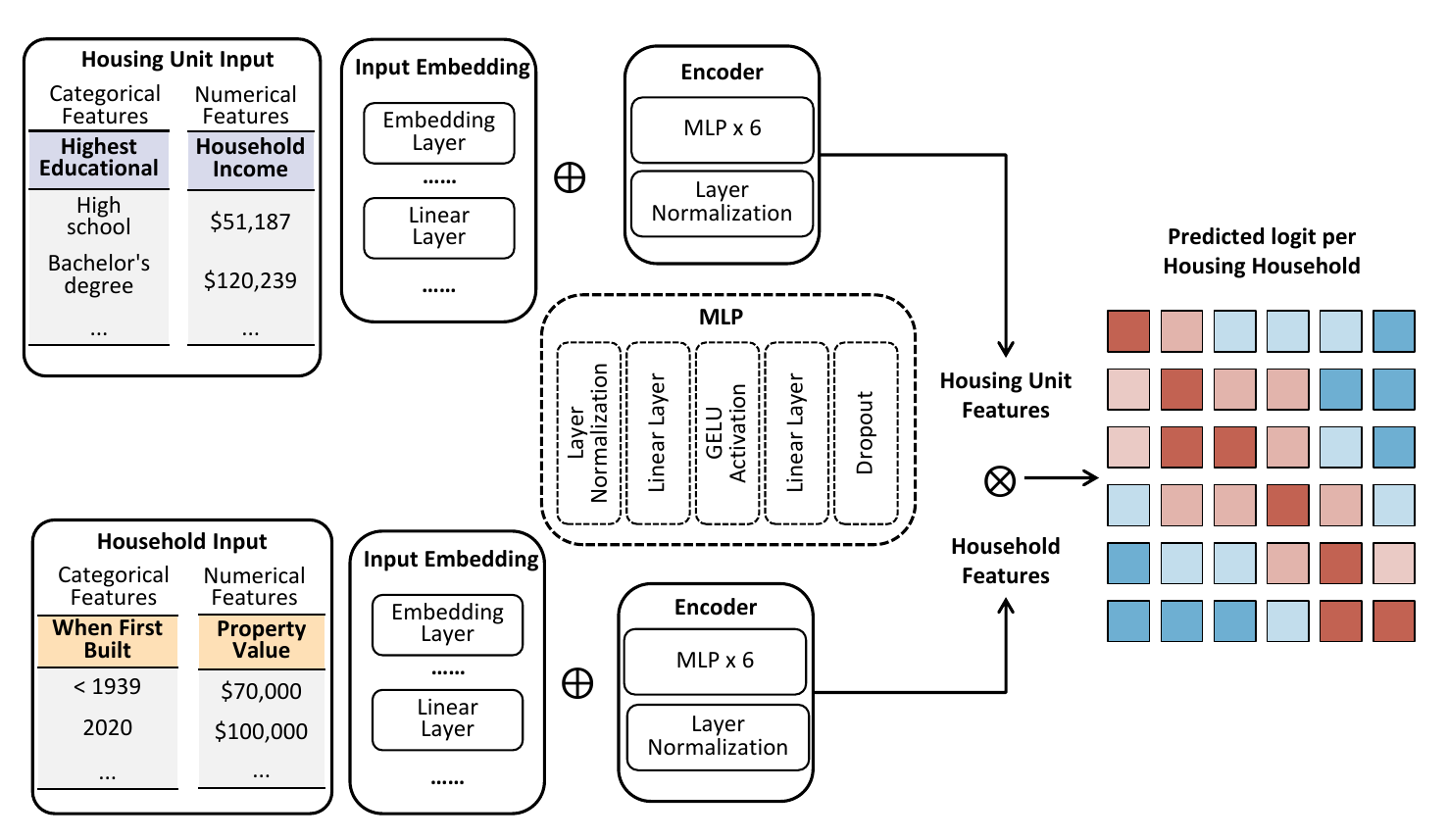}
  \caption{DCL model architecture}
  \label{fig:DL_constrastive_model}
\end{figure}

Common feature alignment tasks, such as image-text pairing in models like CLIP \cite{radford2021learning} and ALIGN \cite{jia2021scaling}, involve matching images of a subject with corresponding text descriptions. In these tasks, although the data is in different modalities (e.g., images and text), both describe the same underlying subject. In our case, however, household sociodemographics describe people’s characteristics, while housing unit features reflect aspects of their living environment, making them semantically and numerically different. Therefore, using a single encoder to process both household and housing data would be ineffective. Instead, we adopt a dual encoder structure, as illustrated in Figure \ref{fig:DL_constrastive_model}. 

This architecture offers several advantages. First, it enables separate feature embeddings for household and housing attributes, allowing each encoder to capture domain-specific representations. Additionally, it supports more efficient and scalable feature matching. By processing housing and household features through two separate encoders (Figure \ref{fig:DL_constrastive_model}) and projecting them into a shared space, our approach avoids the computationally intensive task of computing pairwise similarities across all housing-household combinations, which scales quadratically with sample size. Instead, it reduces complexity to linear, significantly improving scalability.

Each encoder consists of a series of custom Multi-Layer Perceptron (MLP) layers that progressively transform the input features, capturing increasingly abstract representations at each stage. Each MLP layer contains an input layer normalization, a linear transformation, a GELU activation, a second linear transformation, and a dropout. After encoding, the housing and household features are projected into a shared latent space, where a dot product operation computes their similarity. The resulting value indicates the likelihood of a match.

The pseudo-labels we generated—including assumed positive pairs from clustering and assumed negative pairs from outside clusters—introduce noise into the contrastive learning environment. When the model encounters these noisy labels, it adjusts its parameters to fit potentially inaccurate signals, leading to frequent and large parameter fluctuations that can compromise model stability and generalization. To address this, we incorporate momentum distillation \cite{chen2020simple, li2021align}, a training technique that uses a moving average to update model parameters. This approach ensures that model weights evolve gradually, helping to prevent abrupt fluctuations during training with noisy data.

The loss function plays a crucial role in accurately learning housing-household relationships. While the conventional SoftMax-based contrastive loss, typically implemented as InfoNCE (Information Noise-Contrastive Estimation), is effective in single-positive tasks like image-text pairing, it falls short for our many-to-many matching task, where multiple valid matches may exist for each household or housing unit. In this case, InfoNCE’s global normalization does not apply to our context as it assumes a single positive match per query. Additionally, its batch-wide normalization and numerical stabilization techniques (maximum logit subtraction) create significant computational overhead, which poses efficiency challenges for large-scale housing datasets \cite{zhai2023sigmoid}. 

The sigmoid loss function addresses these challenges and has shown effectiveness in multi-modal contrastive learning \cite{zhai2023sigmoid}. First, its formulation inherently supports the many-to-many nature of housing-household pairings. Second, it removes the need for batch-wide normalization as required in InfoNCE, allowing for more efficient processing of large housing datasets. Moreover, the sigmoid function yields more stable gradients for both positive and negative samples, which is critical for learning robust representations in noisy and label-scarce settings. Therefore, we adopt the sigmoid-based loss function, defined as:

\begin{equation} \mathcal{L}_{ij} = \log\frac{1}{1+e^{z_{ij}(-t \cdot \mathbf{h}_i\cdot\mathbf{u}_j+b)}} \end{equation}

\begin{equation} \mathcal{L} = -\frac{1}{|\mathcal{B}|}\sum_{i=1}^{|\mathcal{B}|}\sum_{j=1}^{|\mathcal{B}|}\mathcal{L}_{ij} \end{equation}

where $\mathbf{h}_i$ and $\mathbf{u}_j$ represent the embeddings of the $i$-th household and $j$-th housing unit, respectively. $z_{ij}$ is the binary indicator representing whether they form a positive housing-household pair based on the training set. $t$ is the learnable temperature that scales the dot product similarity, enabling the model to adjust its sensitivity to variations in housing unit and household features—a critical factor given the diversity within the microdata. $b$ serves as a dynamic threshold for determining housing-household matches, helping to address the class imbalance in the training set.

\section{Performance Evaluation Metrics}\label{sec:performance-evaluation}

We assess the learned housing-household relationship by using it to match housing units with households and measure the accuracy of these matches. To capture various dimensions of matching performance, we employ multiple evaluation metrics. 

\textbf{Average Precision (AP)} \cite{davis2006relationship} is used to evaluate the quality of housing-household matching predictions, especially the extent to which the model ranks the positive ones higher as desired. Our proposed DCL pipeline generates a matching score for each housing unit-household pair. Depending on the threshold selected (above which a pair is classified as a positive match and below which it is considered a negative match), different matching results are produced. By plotting precision against recall across varying thresholds, we generate a Precision-Recall (PR) curve. AP is computed as the area under this curve, ranging from 0 to 1, defined as:
\begin{equation}
  AP = \sum_{n} (R_n - R_{n-1}) P_n
\end{equation}
where $P_n$ and $R_n$ are the precision and recall at the $n$-th threshold, respectively. A higher AP indicates superior model performance, reflecting an optimal balance between precision and recall across the entire range of thresholds.

\textbf{Normalized Discounted Cumulative Gain (NDCG)} \cite{jarvelin2002cumulated} is adopted to assess how well the model prioritizes the most suitable housing units for each household. In this problem, while multiple housing units may be suitable for a household, they vary in compatibility. The model predicts a ranked list of matching housing units for each household, with scores indicating the strength of each match. NDCG evaluates whether the highest-ranked units are truly the best matches, effectively capturing such "degree of matching". A high NDCG score indicates the model effectively captures the relationship between household and housing unit features, prioritizing the most relevant matches. The calculation of NDCG at position $k$ is defined as:

\begin{equation}
NDCG@k = \frac{DCG@k}{IDCG@k} = \frac{\sum_{i=1}^k \frac{y_{true}[\pi_i]}{\log_2(i + 1)}}{\sum_{i=1}^k \frac{y_{true}^*[i]}{\log_2(i + 1)}} 
\end{equation}
where $\pi = \text{argsort}(y_{score})[::-1]$ (descending order indices). $\pi_i$ is the original index at position $i$ after sorting the predictions in descending order, $y_{score}$ represents the predicted matching scores from the model. $y_{true}^* = \text{sort}(y_{true}$, descending=True) (sorted ground truth). $y_{true}$ represents the actual relevance scores (1 for suitable matches, 0 for unsuitable ones), and $y_{true}^*$ represents sorted ground truth relevance scores in descending order. $k$ refers to the number of top positions to evaluate. The numerator ($NDCG@k$) measures the discounted cumulative gain based on the ranking induced by model predictions. The denominator ($IDCG@k$) computes the ideal discounted cumulative gain from the best possible ranking. The logarithmic discount factor $\log_2(i + 1)$ penalizes relevance scores at lower ranks. Finally, the ratio between DCG and IDCG normalizes the score to $[0,1]$, where 1 indicates perfect ranking and 0 indicates the worst possible ranking.

We compare our approach with state-of-the-art tabular data feature alignment methods, SubTab \cite{ucar2021subtab}, to demonstrate the superior performance of our DCL framework in modeling housing-household relationships. SubTab is a self-supervised learning framework designed to generate meaningful representations for tabular data by dividing features into subsets and reconstructing the original data. To adapt SubTab for our matching task, we made several key modifications to align the method with our data's characteristics. First, we applied one-hot encoding to preprocess categorical features, enabling the model to handle the diverse categorical variables in the dataset. Second, instead of randomly dividing features as in the original SubTab, we partitioned the data into household and housing unit sub-tables. Third, to account for differences in feature set sizes, we padded the sub-tables to ensure uniform feature dimensions. After training the encoder-decoder architecture, we extracted feature representations from the encoders for both households and housing units. Using the outer product of these representations, we computed the logits for the matching relationships, following the same process as in our proposed method.

\section{Performance Evaluation of DCL in a Controlled Experiment} \label{sec:theoretical_performance}

We designed our DCL framework to be dataset-agnostic, making it applicable to any matching tasks between two sets of features. Ideally, this method should be evaluated using a ground truth dataset, where both positive and negative matches, along with their matching degrees, are explicitly known. However, in real-world scenarios, such a dataset is often unavailable, posing a limitation to direct validation. Many real-world tabular datasets only capture positive co-occurrences. For instance, which individuals live in which households, or which individuals hold which job types. These observed pairings are often limited due to the sample size and may not represent the most optimal matches. For example, a household residing in a particular housing unit does not necessarily indicate that it is the most compatible option, only a feasible one. Additionally, datasets rarely provide explicit non-matching information, such as which individuals should not be living in a given house. To overcome this limitation, we construct a synthetic ground truth dataset where all matching relationships, including both positive and negative cases, are fully known, enabling a more rigorous evaluation of our method’s performance.

\subsection{Synthetic ground truth generation}
\label{sec:synthetic_data}

A synthetic ground truth can be constructed by interlinking two datasets through predefined non-linear relationships. Given a table $\mathbf{X}$ with $N$ records, where each record $\mathbf{x}_i = [x_{i1}, x_{i2}, ..., x_{iM}]$ consists of both categorical and numerical features, a corresponding table $\mathbf{Y}$ can be generated for controlled experiments by applying the following non-linear transformations:
\begin{equation}
  \mathbf{y}_i = f(\mathbf{x}_{i1}, \mathbf{x}_{i2}, ..., \mathbf{x}_{iM}) = f(\alpha_1\mathbf{x}_{i1} + \alpha_2\mathbf{x}_{i2} + ... + \alpha_M\mathbf{x}_{iM})
\end{equation}
where $\mathbf{y}_i$ represents the target feature vector for the $i$-th record, and $\mathbf{x}_{ij}$ denotes the $j$-th feature of the $i$-th record in the input table $ \mathbf{X}$. The coefficient $\alpha_j$ controls the contribution of each feature. The function $f_k(\cdot)$ then applies a non-linear transformation to the weighted sum of these powered features, generating the $k$-th target feature. This formulation constructs complex yet controlled relationships between the two tables, mirroring real-world scenarios where features across datasets are interconnected through intricate non-linear dependencies. 

Specifically, in this experiment, the first table includes the categorical feature $c_i \in {1,2,3,4,5}$, and numerical features $n_{i1}$ and $ n_{i2} \sim \mathcal{U}(0,1)$, drawn from a uniform distribution. The second table consists of three features derived through the following non-linear transformations:

\begin{equation}
\begin{aligned}
y_{i1} &= \sin(\pi(0.5n_{i1} + 0.3n_{i2} + 0.2c_i/5)) \\
y_{i2} &= \exp(0.4n_{i1} + 0.4n_{i2} + 0.2c_i/5) \\
y_{i3} &= \tanh(0.3n_{i1} + 0.3n_{i2} + 0.4c_i/5)
\end{aligned}
\end{equation}

These features first go through weighted, non-linear operations and then apply different non-linear transformations, including trigonometric, exponential, and hyperbolic functions. To ensure a balanced contribution from all features, the categorical variable $c_i$ is normalized by its maximum value of 5. This design preserves interpretability by distinctly separating the weighted non-linear combination from the subsequent transformations while still maintaining complex relationships.

We generate 6,400 unique samples using stratified sampling to ensure a balanced representation across all categorical values. The dataset is then divided into 5,120 samples (80\%) for training, 1,024 (16\%) for validation, and 256 (4\%) for testing. To prevent data leakage and ensure a robust evaluation, we enforce strict partitioning, ensuring that no sample appears in multiple splits while maintaining a balanced categorical feature distribution across all subsets. This synthetic ground truth dataset enables precise performance evaluation by offering known relationships, allowing for a systematic assessment of model capabilities under controlled complexity, free from noise and missing values.

\subsection{Performance evaluation of joint variable relationship learning}

We applied the proposed DCL model to learn the constructed complex joint relationships in the synthetic ground truth dataset. Performance evaluation in Table \ref{tab:synthetic_results} using Average Precision (AP) and Normalized Discounted Cumulative Gain (NDCG) demonstrates the superior effectiveness of our DCL model compared to the state-of-the-art SubTab model. 

\begin{table}[!ht]
  \centering
  \caption{Variable joining performance comparison on synthetic ground truth}
  \label{tab:synthetic_results}
  \begin{tabular}{@{}lcc@{}}
    \toprule
    \textbf{Model} & \textbf{Average Precision (AP)} & \textbf{NDCG} \\
    \midrule
    DCL (Ours) & \textbf{0.8690} & \textbf{0.9791} \\
    SubTab & 0.3044 & 0.8493 \\
    \bottomrule
  \end{tabular}
\end{table}

The DCL model achieved an AP of 0.8690, substantially outperforming SubTab’s 0.3044. The significant improvement in AP indicates that our model more accurately distinguishes between positive and negative matching pairs while maintaining higher precision across varying recall thresholds. Regarding ranking quality, our model attained an NDCG score of 0.9791, compared to SubTab’s 0.8493. This higher NDCG score underscores the DCL model’s ability to rank truly matching pairs more effectively, which is particularly important in applications such as housing-household matching, where ranking quality directly impacts decision-making. 

The superior performance of DCL demonstrates that the contrastive learning framework effectively captures meaningful representations, leading to more accurate matching between related entries. The performance gap between DCL and SubTab can be attributed to its dual-encoder architecture (Figure \ref{fig:DL_constrastive_model}). Unlike SubTab, which relies on a single encoder to reconstruct and represent the entire table, DCL employs a dual-encoder approach to generate distinct representations for each table, resulting in a substantial performance boost. Its strong results on the synthetic ground truth dataset validate its effectiveness, assuring its applicability to real-world scenarios where joint relationships are only partially observed, such as housing-household matching.

\section{Case Study in Delaware and North Carolina}\label{sec:results}

Building on DCL’s strong performance with synthetic ground truth, we applied it to learn the joint housing-household relationship. Delaware was used to evaluate the model’s accuracy in matching housing units with households, while North Carolina served as a test site to assess its transferability. The Delaware microdata consists of 18,641 housing-household co-occurrence records, while North Carolina’s dataset contains 198,037 entries.

\subsection{Microdata for housing-household relationship modeling and testing}

The microdata includes a wide range of variables. For this study, we selected 11 key variables each from the household and housing unit data (Table \ref{tab:selected_variables}), focusing on those assumed to be most relevant for understanding their relationship in our applications. We included tenure attributes for both housing units and households, as households can be renters or owners, and housing units can be rental or owned properties. This selection also enhances the model’s test set. Additional variables can be included as needed to better align with specific user needs and application goals.

\begin{table}[!ht]
  \caption{Selected variables for housing-household relationship modeling}
  \resizebox{\textwidth}{!}{%
  \begin{tabular}{ll|ll}
\hline
\multicolumn{2}{c|}{\textbf{Housing unit}}          & \multicolumn{2}{c}{\textbf{Household}}                                \\ \hline
\textbf{Variable Name} & \textbf{Description}       & \textbf{Variable Name} & \textbf{Description}                         \\ \hline
ACR                    & Lot size                   & NP                     & Number of persons                            \\
BDSP                   & Number of bedrooms         & GRNTP                  & Gross rent                                   \\
BLD                    & Units in structure         & GRPIP                  & Gross rent as percentage of household income \\
MRGP                   & First mortgage payment     & HHL                    & Household language                           \\
RMSP                   & Number of rooms            & HHLDRAGEP              & Age of the householder                       \\
RNTP                   & Monthly rent               & HHLDRRAC1P             & Race of the householder                      \\
TEN\_U                 & Tenure                     & TEN\_H                 & Tenure                            \\
VALP                   & Property value             & HUPAC                  & Household presence and age of children       \\
VEH                    & Vehicles available         & R65                    & Presence of persons 65 years and over        \\
YRBLT                  & When structure first built & SCHL                   & Highest education attainment                 \\
TAXAMT                 & Property taxes             & DIS                    & Number of disabilities                       \\ 
                       &                            & HINCP                    & Household income \\ \hline
\end{tabular}
}
\label{tab:selected_variables}
\end{table}

Our dual-encoder contrastive learning model is designed to adapt to noisy training data. However, it is impractical to test the model on the enhanced microdata, from which the training data is derived, as clustering introduces invalid positive and negative household-housing pairs. To ensure reliable evaluation, we revert to the original microdata for model testing.

However, the lack of ground truth negative pairs in the original microdata limits our ability to fully assess the model’s discriminative capacity. Without negative test data, the model may develop a bias toward positive predictions, leading to inflated performance scores and preventing an accurate evaluation of false positive rates, precision, and recall.

Many existing studies address this challenge by relying on expert input to create their own test sets. For instance, Radford et al. \cite{radford2021learning} manually curated datasets for zero-shot evaluation instead of relying on potentially noisy, web-crawled data. Jia et al. \cite{jia2021scaling} annotated image classification datasets to ensure evaluation reliability. However, manually defining negative housing-household pairs is more complex than image-text labeling, as real-world exceptions are common. For example, a 20-year-old could be a college professor, or a household earning under \$75k might live in a million-dollar home due to inheritance. Given these complexities, we can only apply fundamental rules in housing-household relationships to curate negative pairs for testing.

In this research, we use tenure matching as the primary criterion for identifying false matches. Specifically, we structure the microdata so that tenure acts as a shared attribute between household and housing unit features, where a household can be either a renter or owner, and a housing unit can be rented or owned. If a renter is paired with an owned property, we classify this as a negative housing-household pair. Using the household and housing records in the microdata, we apply this rule to create a set of negative pairs based on mismatched tenure statuses.

We acknowledge that tenure mismatch $\{\mathtt{S}\}$ is only a subset of the constraints $\{\mathtt{C}\}$ defining negative housing-household relationships. However, this rule serves as a practical starting point, especially given the limited understanding of housing-household relationships, which is also the motivation behind this research. To ensure this rule-based subset serves as an effective test set of the model’s performance, we avoid labeling or sampling negative pairs based on tenure status when generating the training data, preventing this tenure constraint leak into the training process. We hope this subset will demonstrate the model's capability, and as more insights into housing-household relationships emerge, we can further refine the test set.

\subsection{Housing-household joining performance in Delaware}\label{sec:hu_matching}

\begin{figure}[!ht]
  \centering
  \includegraphics[width=0.99\textwidth]{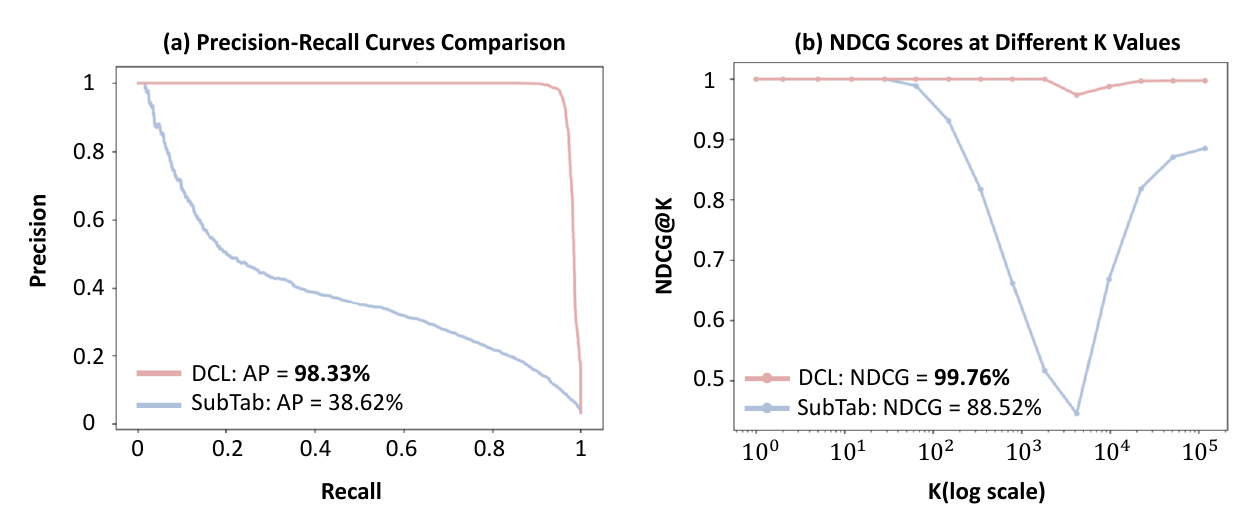}
  \caption{Model performance}
  \label{fig:pr_curves}
\end{figure}

We first tested the model using Delaware's microdata. Households in Delaware microdata were grouped into 3,000 clusters, with an average of 5 households per cluster. Figure \ref{fig:pr_curves}(a) presents the precision-recall curves comparing our DCL approach to the SubTab method. The curves show the superior performance of our method, which maintains high precision (>0.95) across a wide range of recall values. SubTab's precision diminishes rapidly as recall increases, indicating less reliable matching predictions. Overall, our method achieves an AP of 98.33\%, substantially outperforming SubTab's 38.62\%. This result highlights our method's superior ability to reliably and accurately identify positive housing-household matches. Our approach also excels in the NDCG metric (Figure \ref{fig:pr_curves}(b)), achieving 99.76\% compared to SubTab's 88.52\%. This highlights our model's effectiveness in ranking the most compatible housing unit for a given household at the top when calculating matching scores.

The superior performance of our method can be attributed to its tailored architectural design. Unlike SubTab, which is optimized for learning general representations across entire tables, our model is specifically designed to capture representations that determine whether a given household and housing unit form a valid match. This targeted approach enables our DCL model to effectively capture the intricate nuances of housing-household relationships.

\begin{figure}[!ht]
  \centering
  \includegraphics[width=\textwidth]{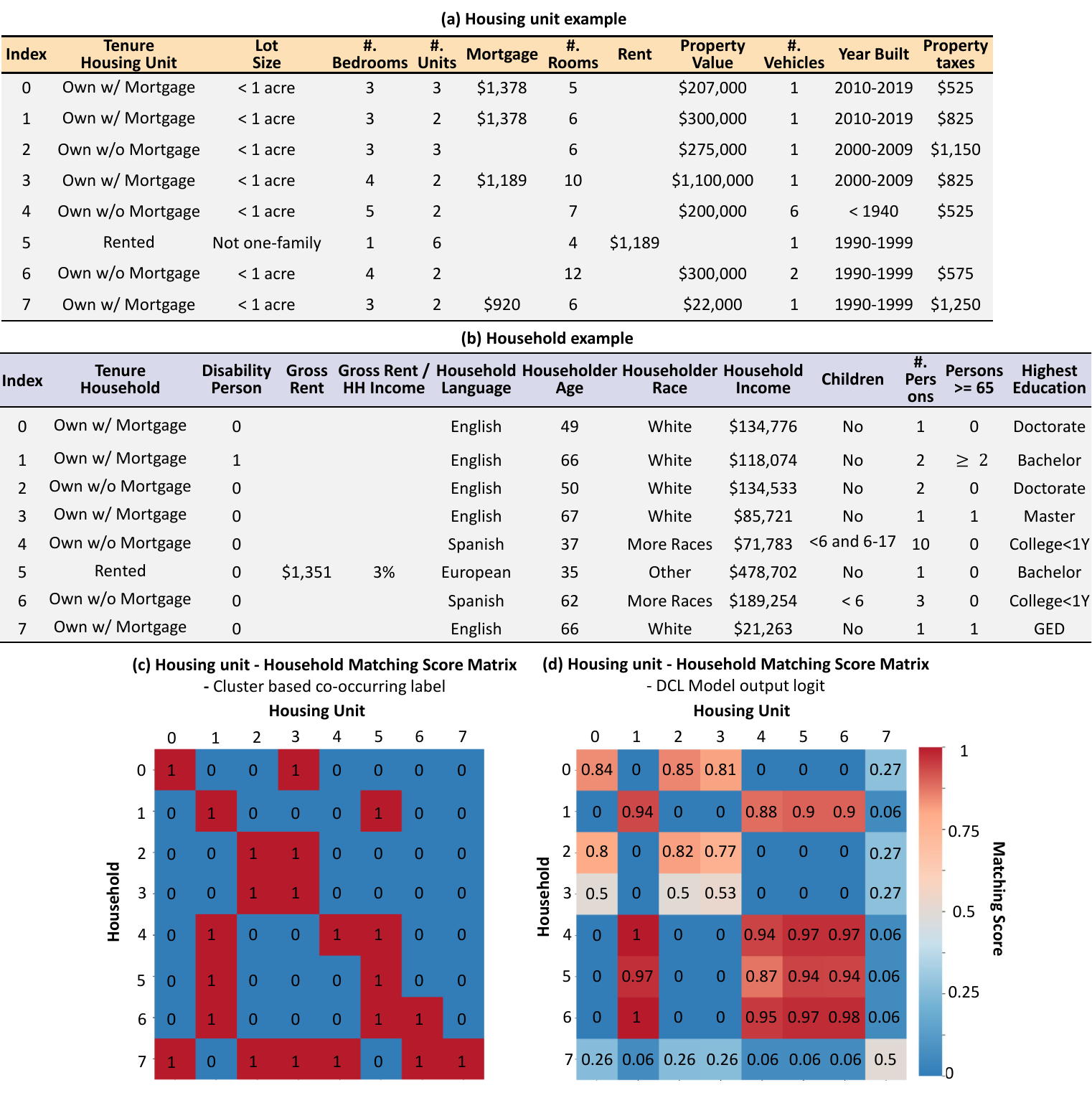}
  \caption{Housing-household matching illustration}
  \label{fig:matching_scores}
\end{figure}

Due to the high dimensionality of the learned housing-household relationship, it is impractical to summarize a rule that can describe which households are likely to match with specific housing units. Instead, we visually present the matching results for a randomly selected set of housing units and households (Figure \ref{fig:matching_scores}(a) and (b)). Figure \ref{fig:matching_scores}(c) illustrates cluster-based housing-household pairs. Households 2 and 3 are grouped into the same cluster as they are both high-income households with mortgages. Correspondingly, their associated housing units, 2 and 3, are also clustered together since they are mortgaged properties with multiple bedrooms. Due to the clustering, Household 2 (3) is also paired with Housing unit 3 (2). Figure \ref{fig:matching_scores}(d) shows our DCL model’s matched and unmatched housing-household pairs. Unlike traditional binary classification, our model assigns a continuous score to each pair, indicating their degree of fitness. This scoring approach provides richer insights compared to binary labels, enabling the model to capture subtle differences in compatibility between households and housing units. Moreover, our model generates a broader set of potential matches beyond those observed in the microdata. As previously discussed, microdata is a sampled dataset that does not capture all possible housing-household pairs. Additionally, the housing unit a household resides in is not necessarily the only viable option. By expanding the scope of possible matches, our model better reflects the broader range of housing choices available to households.

\subsection{Sensitivity analysis of cluster size}
\label{sec:sensitivity}

Clustering is a key step in our data preprocessing, making the choice of the number of clusters a critical factor for our model's performance. To evaluate the effect of cluster quantity on learning outcomes, we performed a sensitivity analysis. Table \ref{tab:cluster_impact} summarizes the model's performance across varying numbers of clusters. The results show that as the number of clusters increases, the model's performance initially improves, but beyond a certain threshold, it begins to decline. This indicates that a finer-grained clustering can better capture and enhance the learning of the housing-household relationship. Notably, even without clustering preprocessing (treating each housing unit and household as its own singleton cluster), our model maintained strong performance. 

\begin{table}[!ht]
  \centering
  \caption{Impact of Different Cluster Numbers on the Performance of Our Method}
  \label{tab:cluster_impact}
  \begin{tabular}{@{}lccccc@{}}
  \toprule
                           & \textbf{15,220} & \textbf{3,000} & \textbf{2,000} & \textbf{1,000} & \textbf{100} \\
                           & \textbf{Clusters} & \textbf{Clusters} & \textbf{Clusters} & \textbf{Clusters} & \textbf{Clusters} \\
                          \textbf{Medium number of} & & & & & \\ 
                          \textbf{households per cluster} & \textbf{Singleton} & 5 & 8 & 16 & 158 \\ \midrule
  \textbf{Average Precision (AP)} & 83.99\%             & \textbf{98.33\%}       & 97.45\%                & 61.68\%                & 43.97\%               \\
  \textbf{NDCG}              & 97.76\%             & \textbf{99.76\%}       & 99.65\%                & 92.48\%                & 89.65\%               \\ \bottomrule
  \end{tabular}
\end{table}

This phenomenon can be attributed to the inherent noise in the self-supervised training dataset. Because microdata only covers a sample (1\%-5\%) of the total population, without clustering, many potential positive housing-household pairs are incorrectly labeled as negative, resulting in a high rate of false negatives. Conversely, using large-scale clustering causes most housing-household pairs to be labeled as positive, significantly increasing false positives. As cluster size grows, the false positive rate escalates rapidly (see Figure \ref{fig:clustering_error}), leading to diminished model performance. By comparing these extreme conditions, we observe that the model handles fine-grained clustering more effectively than coarse clustering. Therefore, we adopt a strategy of small cluster sizes, with each cluster containing approximately 4–8 samples. In our experiment, 3000 clusters yield an excellent performance. This approach minimizes false negatives while controlling the false positive rate, striking a critical balance that preserves high model performance.

\subsection{Transferability test in North Carolina}

To assess our model's generalization capability across different geographic regions, we conducted transfer learning experiments in North Carolina (NC). The NC microdata contains 161,135 households, and based on insights from the Delaware sensitivity analysis, we maintained 3,000 clusters for the NC study, resulting in a cluster size of 53 households.

\begin{figure}[!ht]
  \centering
  \includegraphics[width=0.99\textwidth]{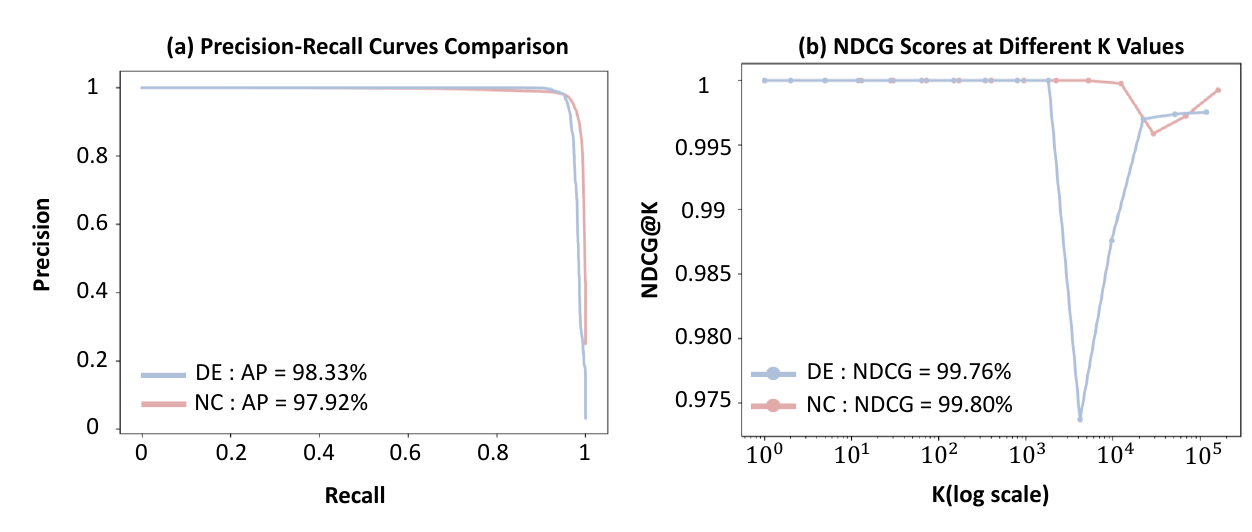}
  \caption{Transferability test in North Carolina (NC)}
  \label{fig:transferability}
\end{figure}

The results (Figure \ref{fig:transferability}) show an outstanding transfer learning performance. Using the model trained on North Carolina microdata, we achieved an Average Precision of 97.92\% and NDCG of 99.80\%. This highlights the robustness of our proposed data preprocessing and DCL pipeline, suggesting that our approach effectively captures key features of housing-household relationships across varying sociodemographic and built environment contexts.

\subsection{Influential factors of housing-household joining}

To gain insights into how our deep learning model makes housing-household matching decisions, we utilize a post-hoc explainable AI technique \cite{gilpin2018explaining}. This approach enhances the interpretability of our trained DCL model without altering its structure or training process. The approach is applied post-training to provide clarity on the model's outputs and the reasoning behind its decisions. Such techniques are particularly valuable for "black-box" models like our deep contrastive model, which, despite their high accuracy, are inherently difficult to interpret. Ensuring interpretability is critical not only for validating the model but also for practical applications in housing policy, where understanding the reasoning behind matching decisions is as important as the decisions themselves.

\begin{figure}[!ht]
  \centering
  \includegraphics[width=0.99\textwidth]{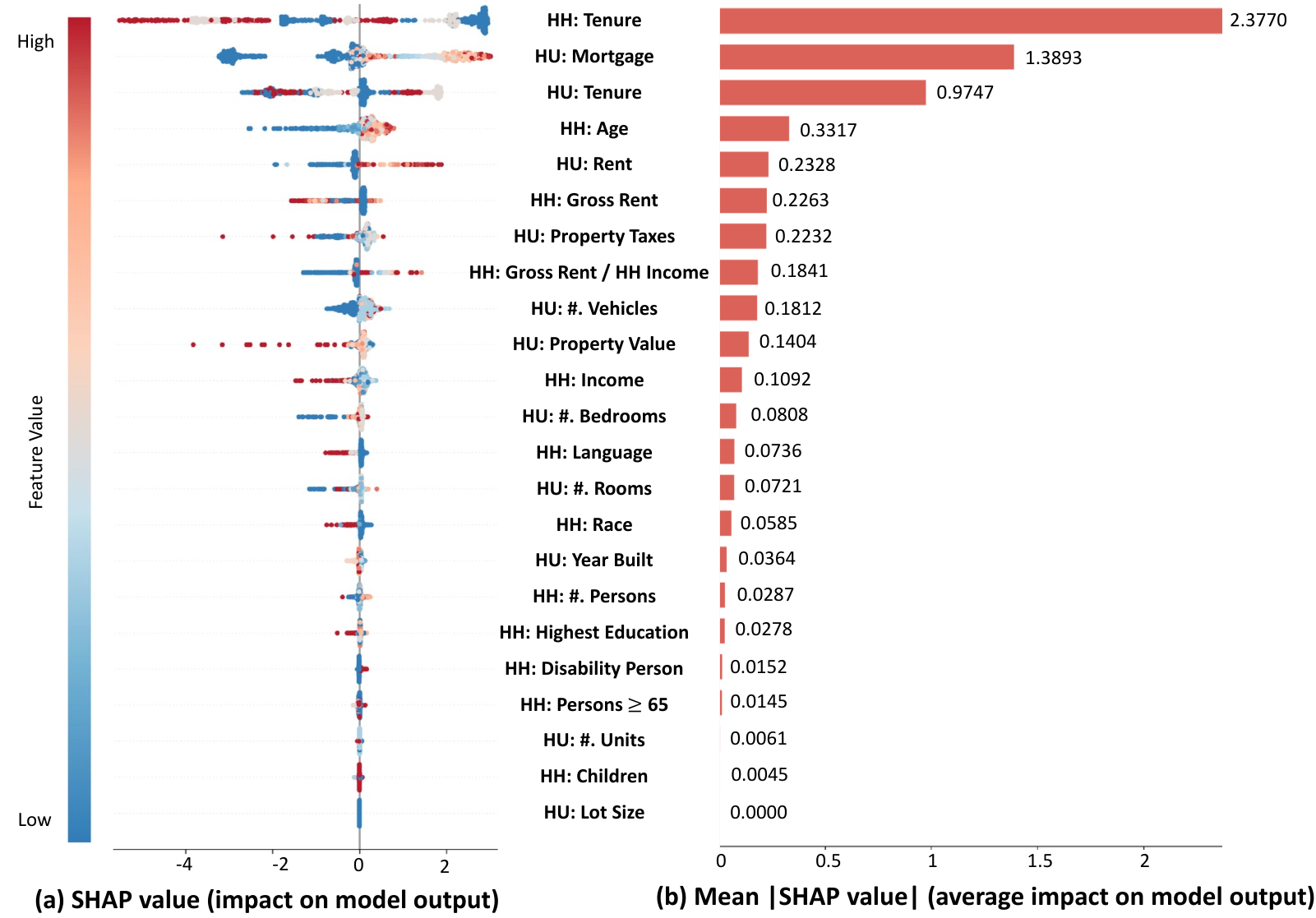}
  \caption{SHAP analysis of feature importance in housing-household matching}
  \label{fig:shap_summary}
\end{figure}

In this study, we utilized an XGBoost model as a post-hoc method to analyze feature contributions within our housing-household unit matching framework. To quantify the impact of each feature on the model's predictions, we employed SHAP (SHapley Additive exPlanations) values, which are widely recognized for their consistency and accuracy in interpreting complex models \cite{lundberg2018consistent}. Figure \ref{fig:shap_summary}(a) displays a beeswarm plot of SHAP values, illustrating the distribution and influence of individual feature values on the model’s predictions. Additionally, Figure \ref{fig:shap_summary}(b) summarizes the mean absolute SHAP values, providing a clear ranking of each feature’s overall importance.

The analysis reveals a distinct hierarchy in feature importance, with tenure-related attributes (HH\_TEN: 2.2141, HU\_TEN: 0.8683) and financial indicators (HU\_MRGP: 1.3854) playing dominant roles in the matching decisions. The significant variation in SHAP values for these key features (e.g., 2.4913 for HH\_TEN) with both positive and negative effects on predictions underscores the complexity of the housing-household matching process, which is highly context-dependent and influenced by the specific values of these features and their interactions with other variables.

More important, the analysis also reveals that physical attributes of housing units, such as the number of bedrooms (HU\_BDSP: 0.0749) and rooms (HU\_RMSP: 0.0640), have relatively modest importance compared to financial factors. This finding extends beyond existing housing-household matching research, which primarily focuses on building capacity and geospatial density, uncovering new key drivers in the alignment between households and housing units.

\section{Conclusion}
This paper presents a deep contrastive learning (DCL) pipeline (Figure \ref{fig:pipeline}) to establish a high-dimensional statistical relationship between housing units and households, addressing a critical gap in disaster impact analysis. The modeling of this relationship is challenging due to several factors: the absence of labeled datasets with embedded matching degrees, limited positive housing-household pairs with binary outcomes, a lack of information on negative housing-household pairings, the multi-modality issue arising from sociodemographic features describing humans versus structural features describing housing units, and the missing many-to-many pairing information between housing and households.

To tackle these challenges, we leverage the natural housing-household pairing available in microdata, employing co-occurrence as a pretext task for modeling the housing-household relationship (Figure \ref{fig:co-occurring}). To address data limitations, we propose a Bisecting K-Means clustering technique to generate essential training data (Figure \ref{fig:cluster} and \ref{fig:clustering_error}). Furthermore, in the absence of a labeled dataset, we implement a self-supervised learning approach—deep contrastive learning (DCL)—with a sigmoid contrastive loss function (Figure \ref{fig:DL_constrastive_model}). This technique enables the model to learn the many-to-many compatibility between housing units and households, producing a matching score that quantifies the degree of alignment between them.

To rigorously evaluate the performance of the proposed DCL variable joining framework, we constructed a synthetic ground truth with a known non-linear relationship between two tables. The controlled experiment demonstrates that our DCL framework can more effectively capture variable relationships compared to the state-of-the-art SubTab method (Table \ref{tab:synthetic_results}). This superior performance confirms its suitability for applying to microdata in housing-household relationship modeling.

The empirical experimental results in Delaware demonstrate that our approach outperforms state-of-the-art models across multiple metrics, including Average Precision (AP) and Normalized Discounted Cumulative Gain (NDCG), confirming its superior ability to identify and rank compatible housing-household pairs (Figure \ref{fig:pr_curves}). The model not only accurately predicts the matching pairs documented in the microdata but also generates additional potential matches (Figure \ref{fig:matching_scores}). Our analysis shows that the model performs most robustly and reliably when the cluster size is small (Table \ref{tab:cluster_impact}). To further validate the approach, we tested the pipeline in North Carolina, where it delivered similarly excellent performance (Figure \ref{fig:transferability}). To enhance the explainability of the proposed DCL model, we developed a post-hoc analysis using an XGBoost model to identify key factors influencing the housing-household matching process (Figure \ref{fig:shap_summary}). The results reveal that the tenure status of both the housing unit and household, along with the housing unit's mortgage information, are the most significant factors in mapping their relationships.

Despite the advancements achieved in this research, several limitations remain. First, the ACS dataset used is relatively small, representing only a 5\% sample of the actual population and housing units within Public Use Microdata Areas. This limitation may lead to gaps in geographic coverage in certain states, potentially introducing biases that affect the method's alignment with real-world scenarios. Future research could address this issue by integrating data from more diverse sources to improve dataset representativeness. Additionally, while our method for generating negative samples in the test set is innovative, it captures only a subset of the true negative instances. Future studies could investigate more sophisticated approaches to constructing a comprehensive test set, such as developing complex housing-household unmatching rules or engaging human experts to create a large-scale validation dataset. This enhancement would allow for a more robust evaluation of the model's performance.

This study represents a significant leap in housing-household matching by integrating innovative data preprocessing techniques and the DCL method. By moving beyond traditional binary allocation approaches, our method provides a nuanced degree of matching that transcends the basic physical and distributional constraints such as building capacity, population density, and sociodemographic marginal distributions. These advancements offer a more accurate representation of household distributions within the built environment, enhancing our understanding of urban dynamics and social interactions. The improved accuracy has broad implications, supporting more informed urban planning decisions, strengthening disaster response efforts, and deepening our understanding of societal structures within urban environments.

\section*{Acknowledgments}
Shangjia Dong would like to acknowledge funding support from the National Science Foundation \#2443784. Rachel Davison would like to acknowledge funding support from the National Science Foundation \#2209190. Any opinions, conclusions, and recommendations expressed in this research are those of the authors and do not necessarily reflect the view of the funding agencies. The authors would also like to thank the editor and the anonymous reviewers for their constructive comments and valuable insights to improve the quality of the article. 

\bibliographystyle{plainnat}

\label{sec:refs}

\end{document}